\documentclass{article}

\usepackage{arxiv}

\usepackage{newtxtext}
\usepackage{newtxmath}

\usepackage[utf8]{inputenc} 
\usepackage[T1]{fontenc}    
\usepackage[russian,english]{babel} 
\usepackage{hyperref}       
\usepackage{url}            
\usepackage{booktabs}       
\usepackage{amsfonts}       
\usepackage{nicefrac}       
\usepackage{microtype}      
\usepackage{cleveref}       
\usepackage{lipsum}         
\usepackage{graphicx}
\usepackage{natbib}
\usepackage{doi}

\newif\ifuniqueAffiliation
\uniqueAffiliationtrue  

\title{Benchmarking POS Tagging for the Tajik Language: A Comparative Study of Neural Architectures on the TajPersParallel Corpus}

\ifuniqueAffiliation
    \author{%
        \href{https://orcid.org/0000-0003-2525-1183}{\includegraphics[scale=0.06]{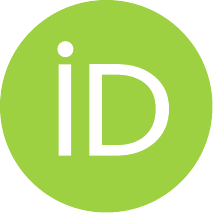}\hspace{1mm}M. K. Arabov}\thanks{Email: \texttt{MKArabov@kpfu.ru}} \\
        Institute of Computational Mathematics and Information Technologies\\
        Kazan Federal University\\
        Kazan, Russia \\
        \texttt{MKArabov@kpfu.ru}
    }
\else
    \usepackage{authblk}
    
    \newbox{\orcid}\sbox{\orcid}{\includegraphics[scale=0.06]{orcid.pdf}}
    \author[1]{%
        \href{https://orcid.org/0000-0003-2525-1183}{\usebox{\orcid}\hspace{1mm}M. K. Arabov\thanks{\texttt{MKArabov@kpfu.ru}}}%
    }
    \affil[1]{Institute of Computational Mathematics and Information Technologies, Kazan Federal University, Kazan, Russia}
\fi

\renewcommand{\shorttitle}{POS Tagging Benchmark for Tajik}

\hypersetup{
    pdftitle={Benchmarking POS Tagging for the Tajik Language},
    pdfsubject={cs.CL},
    pdfauthor={M. K. Arabov},
    pdfkeywords={Tajik language, POS tagging, part-of-speech tagging, low-resource languages, multilingual models, BERT, LoRA, BiLSTM-CRF, benchmark, TajPersParallel},
}

\begin{document}
\maketitle
\begin{abstract}
This paper presents the first benchmark for the task of automatic part-of-speech (POS) tagging for the Tajik language. Despite the existence of multilingual language models demonstrating high effectiveness for many of the world's languages, their capacity for grammatical analysis of Tajik has remained unexplored until now. The aim of this study is to fill this gap through a systematic comparison of classical neural network architectures and modern multilingual transformers.

Experiments were conducted on the TajPersParallel corpus, a parallel lexical resource comprising approximately 44,000 dictionary entries. Due to the absence of full-fledged example sentences in the current version of the corpus, the task was performed at the level of isolated lexical units, representing a challenging case of context-independent classification. The study compares the following architectures: a recurrent BiLSTM-CRF model, as well as multilingual models XLM-RoBERTa (large) \citep{conneau2020_xlmr}, mBERT \citep{devlin2019bert}, ParsBERT (Persian) \citep{farahani2021_parsbert}, and ruBERT (Russian) \citep{kuratov2019_rubert}, adapted using the parameter-efficient fine-tuning method LoRA \citep{hu2022_lora}.

The testing results showed that the best performance is achieved by the mBERT + LoRA model (macro F1-score = 0.11, weighted F1-score = 0.62). It was established that in the absence of syntactic context, all models experience significant difficulty in resolving morphological ambiguity, successfully classifying primarily high-frequency classes (``noun,'' ``adjective'') while demonstrating zero effectiveness for rare function words. Zero-shot evaluation revealed the greatest typological proximity of Tajik to Persian (ParsBERT) and Russian (ruBERT). The obtained results form a foundation for further research and development in the field of automatic processing of the Tajik language.
\end{abstract}


\section{Introduction}
The Tajik language, the official state language of the Republic of Tajikistan and a member of the southwestern group of Iranian languages, occupies a distinct position within the Persian linguistic continuum \citep{oranskii1988}. Unlike Persian in Iran (Farsi) and Dari in Afghanistan, which utilize the Arabic script, Tajik has officially employed the Cyrillic alphabet since 1940. This circumstance, combined with prolonged sociolinguistic isolation, has led to a paradoxical situation: while maintaining deep structural and lexical proximity to Persian \citep{oranskii1988}, Tajik has been almost entirely excluded from the sphere of modern Natural Language Processing (NLP) technologies. As noted in an analytical review of the current state of resources \citep{arabov2025_developing}, the Tajik language is characterized by an acute shortage of basic linguistic tools and annotated corpora, making it one of the least resourced languages in the era of large language model dominance.

A fundamental task, without which the construction of higher-level systems—such as syntactic and semantic analyzers, information extraction systems, and machine translation—is impossible, is automatic part-of-speech (POS) tagging. To date, there exists no standard reference corpus with morphological annotation, no trained POS taggers, and no published benchmarks for the Tajik language that would allow for an assessment of the applicability of modern multilingual models to this linguistic material. Meanwhile, a whole range of specialized NLP libraries has been developed for the closely related Persian language \citep{jafari2025_dadmatools_v2}, making the absence of analogous solutions for Tajik particularly conspicuous.

The emergence of the TajPersParallel Tajik-Persian parallel lexical corpus \citep{arabov2026_tajperslexon}, comprising over 43,000 dictionary entries annotated with parts of speech, creates, for the first time, an empirical foundation for conducting systematic research in the field of Tajik morphology. Furthermore, the widespread adoption of parameter-efficient fine-tuning methods for large language models, specifically Low-Rank Adaptation (LoRA) \citep{hu2022_lora}, opens up the possibility of adapting powerful multilingual transformers to a new language with minimal computational cost. However, the question of how successfully models pre-trained on dozens of languages, including Persian, can distinguish parts of speech in Tajik in the absence of syntactic context remains uninvestigated. The present work aims to fill this gap: its objective is to create the first reproducible POS tagging benchmark for the Tajik language based on the TajPersParallel corpus and to conduct a comparative analysis of the effectiveness of the classical BiLSTM-CRF architecture versus modern multilingual transformers with LoRA adaptation.

\section{Related Work}
This research lies at the intersection of several scientific directions, including the development of NLP tools for the Persian language, the task of transliteration between Tajik and Persian writing systems, and methods for efficient knowledge transfer in low-resource language settings.

In the domain of developing NLP tools for Persian, significant progress has been achieved in recent years. The DadmaTools V2 toolkit \citep{jafari2025_dadmatools_v2} deserves particular attention, as its architecture is based on the use of adapters, reflecting current trends in the modular construction of NLP systems. Other notable developments for Persian include the original DadmaTools toolkit \citep{etezadi2022_dadmatools} and the multifunctional APARSIN benchmark for Iranic languages \citep{jafari2026_aparsin}. However, all of these developments are exclusively oriented toward the Persian language in the Arabic script and cannot be directly applied to Tajik texts in Cyrillic without prior script conversion or targeted model adaptation.

The issue of mutual conversion between the graphic systems of Tajik and Persian has its own research history. One of the first practical transliteration systems, based on statistical machine translation, was proposed by Davis \citep{davis2012_tajik_farsi}. The theoretical foundations for automated conversion were laid in the dissertation research of L.~A.~Grashchenko \citep{grashchenko2003}. In recent years, with the advancement of neural network approaches, more sophisticated solutions have emerged, including the digraphic ParsText corpus \citep{merchant2024_parstext}, as well as models based on the Transformer architecture \citep{sadraeijavaheri2024_transformers, merchant2026_parstranslit}. A comprehensive empirical comparison of transliteration models for the Tajik-Farsi pair, spanning from rule-based to Transformer-based approaches, is presented in \citep{arabov2026systematicbenchmark}, further confirming the close lexical proximity of the two languages. The recent work by the author \citep{kurbonovich2026_character} also demonstrated the high efficiency of a character-level Transformer for this task using the same TajPersParallel corpus.

The key resource enabling the formulation of the POS tagging task for Tajik is the TajPersParallel parallel lexical corpus, described in detail in \citep{arabov2026_tajperslexon}. The corpus includes 43,819 dictionary entries, each containing a Tajik word in Cyrillic script, its Persian equivalent in Arabic script, a part-of-speech tag, and, for some entries, illustrative usage examples. Statistical analysis of the corpus, performed in \citep{arabov2026_tajperslexon} and \citep{arabov2025_developing}, revealed a significant class imbalance: over 90\% of all annotations correspond to nouns and adjectives, which presents specific challenges for classifier training and must be considered when interpreting results. Previously, this corpus was successfully used by the author for transliteration tasks \citep{kurbonovich2026_character, arabov2026systematicbenchmark}; in the present study, however, it is applied for the first time for its direct intended purpose—the construction and evaluation of a POS tagger.

From a methodological standpoint, modern POS tagging for low-resource languages typically relies on pre-trained multilingual models. The Transformer architecture \citep{vaswani2017attention}, built upon the concept of attention mechanisms introduced in \citep{bahdanau2015_nmt}, has become the de facto standard in natural language processing. For adapting large pre-trained models to specific tasks with minimal computational overhead, the LoRA method \citep{hu2022_lora} has become widespread, allowing fine-tuning of only a small number of additional parameters while keeping the main body of the model frozen. This approach has proven effective for dozens of low-resource languages; however, its applicability to the task of Tajik POS tagging has not yet been subjected to experimental verification. As a baseline architecture for comparison, this study employs the BiLSTM-CRF model, which has long dominated sequence labeling tasks and continues to serve as a reliable baseline solution in the absence of large pre-trained embeddings for the target language.

Thus, the distinctive feature of the present work is that it provides, for the first time, a systematic comparison of multiple neural network architectures for Tajik POS tagging, applies the LoRA adaptation method for multilingual transformers to this language for the first time, and publishes reproducible results that establish a reference benchmark for future research.

\section{Data}
\label{sec:data}
The empirical basis for the study is the TajPersParallel Tajik-Persian parallel lexical corpus \citep{arabov2026_tajperslexon, tajiknlpworld_hf}, created and maintained by the author of this work. The corpus contains 43,819 dictionary entries, each comprising the following fields: a Tajik word in Cyrillic script (\texttt{tajik}), its Persian equivalent in Arabic script (\texttt{persian}), a part-of-speech tag (\texttt{part\_of\_speech}), and, optionally, a list of usage examples (\texttt{examples}) extracted from classical and modern Tajik-Persian literature. The total number of unique Tajik words in the corpus is 43,162 lexemes.

The corpus's part-of-speech tagging system comprises 12 classes, listed in descending order of frequency: \textit{{\fontencoding{T2A}\selectfont исм}} (noun, 55.2\%), \textit{{\fontencoding{T2A}\selectfont сифат}} (adjective, 35.6\%), \textit{{\fontencoding{T2A}\selectfont зарф}} (adverb, 3.5\%), \textit{{\fontencoding{T2A}\selectfont феъл}} (verb, 3.2\%), \textit{{\fontencoding{T2A}\selectfont исми хос}} (proper noun, 1.0\%), \textit{{\fontencoding{T2A}\selectfont нидо}} (interjection, 0.6\%), \textit{{\fontencoding{T2A}\selectfont шумора}} (numeral, 0.3\%), \textit{{\fontencoding{T2A}\selectfont пайвандак}} (conjunction, 0.2\%), \textit{{\fontencoding{T2A}\selectfont ҷонишин}} (pronoun, 0.09\%), \textit{{\fontencoding{T2A}\selectfont ҳиссача}} (particle, 0.07\%), \textit{{\fontencoding{T2A}\selectfont пешоянд}} (preposition, 0.06\%), and \textit{{\fontencoding{T2A}\selectfont пасоянд}} (postposition, 0.02\%). As evident from the presented distribution, the corpus is characterized by significant class imbalance: over 90\% of all annotations are concentrated in the two most frequent categories, which is typical for lexicographic resources focused on content words and creates additional difficulties when building classifiers capable of recognizing rare function-word classes.

An important methodological feature of the present study is that the \texttt{examples} field, containing illustrative sentences, is populated for only a subset of entries in the current version of the corpus (22,635 examples), and these examples do not form a coherent corpus of sentences with sequential annotation. Consequently, the POS tagging task in this work is addressed at the level of isolated lexical units; that is, it is formulated as a context-independent classification task of assigning a word to a part-of-speech class based solely on information derived from the word itself and its internal morphemic structure. Although this formulation does not allow for modeling syntactic ambiguity phenomena, which require contextual resolution, it does provide an initial objective assessment of the applicability of modern neural network architectures to the morphological analysis of the Tajik language and establishes a lower bound on the quality achievable without the use of syntactic information.

For the experiments, the corpus was partitioned into training, validation, and test sets in a ratio of 80\% / 10\% / 10\%, preserving the original class distribution (stratified splitting). The final sizes of the sets were: training — 34,529 examples, validation — 4,316 examples, test — 4,317 examples.

\section{Methodology}
This study conducts a comparative analysis of five neural network architectures, encompassing both classical approaches to sequence labeling and modern methods based on pre-trained multilingual transformers with efficient adaptation.

\subsection{Baseline Approach: BiLSTM-CRF}
As a baseline architecture, a bidirectional Long Short-Term Memory network with a Conditional Random Field (CRF) output layer was selected. This architecture has for many years served as the de facto standard for sequence labeling tasks, including POS tagging and named entity recognition, and continues to provide a reliable reference point for evaluating more complex models. The model's input layer consists of a trainable word embedding matrix of dimension 128, initialized randomly, as no pre-trained embeddings exist for the Tajik language. The embedding layer is followed by a bidirectional LSTM with a hidden dimension of 128 in each direction, after which a linear projection layer maps the LSTM outputs to a space of dimension equal to the number of part-of-speech classes. A CRF layer operating on top of the linear outputs performs global normalization and accounts for dependencies between adjacent labels. Since in the present task formulation each sentence consists of exactly one word, the advantages of the CRF in modeling label transitions are not utilized, and the model effectively reduces to a BiLSTM classifier with a cross-entropy loss function.
\subsection{Multilingual Transformers with LoRA Adaptation}
The primary focus of this study is to evaluate the effectiveness of knowledge transfer from multilingual pre-trained models to the task of Tajik POS tagging. To this end, four models were selected, differing in the volume and composition of languages included in pre-training: XLM-RoBERTa-large \citep{conneau2020_xlmr}, trained on 100 languages and considered one of the most powerful cross-lingual models; mBERT (bert-base-multilingual-cased) \citep{devlin2019bert}, covering 104 languages including Persian but not Tajik; ParsBERT \citep{farahani2021_parsbert}, a specialized model for Persian trained on Farsi corpora; and ruBERT \citep{kuratov2019_rubert}, a model for Russian included in the comparison due to the shared Cyrillic script and the historical role of Russian as a language of education and scientific communication in Tajikistan. The choice of ParsBERT and ruBERT is motivated by the hypothesis that models pre-trained on languages closely related to Tajik linguistically (Persian) or sociolinguistically (Russian) may exhibit better knowledge transfer performance compared to universal multilingual models.

For adapting each of the listed models to the part-of-speech classification task, the LoRA (Low-Rank Adaptation) method \citep{hu2022_lora} was employed, which allows fine-tuning only a small number of additional parameters without altering the original model weights. The experiments utilized rank decomposition with rank $r = 8$ and scaling factor $\alpha = 16$, applied to the query and value matrices within the self-attention blocks. This approach enables the training of models with billions of parameters on limited computational resources, which is particularly relevant for research in the field of low-resource languages. All models were fine-tuned for three epochs with a batch size of 16, using the AdamW optimizer and a learning rate of $2 \times 10^{-5}$, with a linear learning rate schedule and an initial warm-up phase.

\subsection{Evaluation Metrics}
The primary metrics selected for model comparison are accuracy, as well as macro and weighted F1-scores. Macro-F1, calculated as the arithmetic mean of F1-scores across all classes, is especially important under conditions of severe class imbalance, as it prevents the high frequency of dominant classes from masking a model's complete inability to recognize rare categories. Weighted F1-score, conversely, accounts for class frequency and reflects the practical applicability of the model in scenarios where correct identification of frequent words is more important than recognition of rare ones. For each model configuration, the experiment was repeated with two different pseudo-random number generator seeds, allowing for an assessment of result stability and the calculation of standard deviations for the metrics.

\subsection{Zero-shot Evaluation}
Additionally, for each of the transformer models, an evaluation was conducted in a zero-shot setting, wherein the model is applied to the test set without any fine-tuning on Tajik data. In this setting, the classification layer, initialized randomly, remains untrained, and the model can rely solely on the internal representations formed during pre-training. The results of the zero-shot evaluation serve as an indicator of the degree of typological proximity between Tajik and the languages on which the model was trained.

\section{Results}
This section presents and analyzes the experimental results obtained from the comparative evaluation of five neural network architectures on the task of context-independent part-of-speech tagging for the Tajik language using the TajPersParallel corpus. The analysis covers overall quality metrics, zero-shot evaluation results, per-class performance, and error analysis.

\subsection{Overall Model Comparison}
Table~\ref{tab:main_results} presents the final accuracy, macro-averaged F1-score (Macro F1), and weighted F1-score (Weighted F1) for each of the considered models on the test set. The data are based on the final results saved in \texttt{test\_summary.csv}.

\begin{table}[htbp]
\centering
\caption{Comparison of POS tagging model performance on the test set}
\label{tab:main_results}
\begin{tabular}{lccc}
\toprule
\textbf{Model} & \textbf{Accuracy} & \textbf{Macro F1} & \textbf{Weighted F1} \\
\midrule
BiLSTM-CRF & 0.551 & 0.059 & 0.392 \\
XLM-RoBERTa + LoRA & 0.616 & 0.098 & 0.563 \\
mBERT + LoRA & 0.651 & 0.111 & 0.618 \\
ParsBERT + LoRA & 0.601 & 0.099 & 0.561 \\
ruBERT + LoRA & 0.614 & 0.101 & 0.572 \\
\bottomrule
\end{tabular}
\end{table}

\begin{figure}[htbp]
    \centering
    \includegraphics[width=0.8\linewidth]{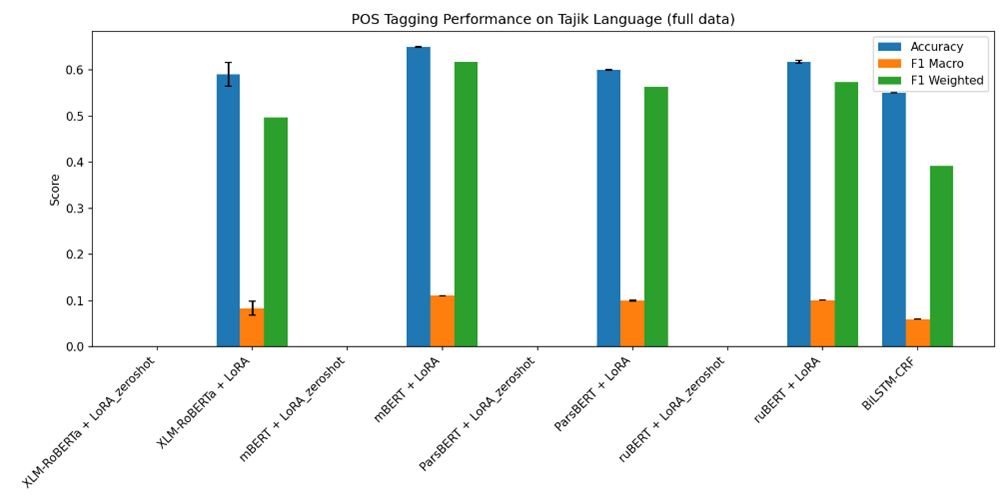}
    \caption{Comparative visualization of quality metrics for POS tagging models on the test set (Accuracy, Macro F1, Weighted F1).}
    \label{fig:comparison}
\end{figure}

As seen from Table~\ref{tab:main_results}, all models based on pre-trained transformers with LoRA adaptation outperform the baseline BiLSTM-CRF architecture across all three metrics. The best performance is achieved by the mBERT + LoRA model, attaining an accuracy of 0.651, macro F1 of 0.111, and weighted F1 of 0.618. It is followed closely by ruBERT + LoRA and ParsBERT + LoRA, which exhibit similar weighted F1 values (0.572 and 0.561, respectively). The XLM-RoBERTa-large model, despite its largest size and broad language coverage, performs comparably to ruBERT but exhibits noticeably higher instability between runs: the standard deviation of accuracy for XLM-RoBERTa was 0.025, whereas for the other models it did not exceed 0.003. A visual comparison of all investigated architectures is provided in Figure~\ref{fig:comparison}.

As illustrated in Figure~\ref{fig:comparison}, the application of LoRA adaptation to multilingual transformers yields a statistically significant increase in all target metrics compared to the classical BiLSTM-CRF architecture. The gap in weighted F1-score is substantial: mBERT + LoRA outperforms the baseline by nearly 23 percentage points. At the same time, the high accuracy (Accuracy $> 0.60$) against the backdrop of extremely low macro F1 ($< 0.12$) indicates that the models optimize their weights primarily for the dominant classes, neglecting rare categories. This pattern is a direct consequence of the severe class imbalance in the training set (detailed in Section~\ref{sec:data}) and the fundamental absence of syntactic context necessary for resolving morphological ambiguity.

\subsection{Zero-shot Evaluation}
\begin{figure}[htbp]
    \centering
    \includegraphics[width=0.8\linewidth]{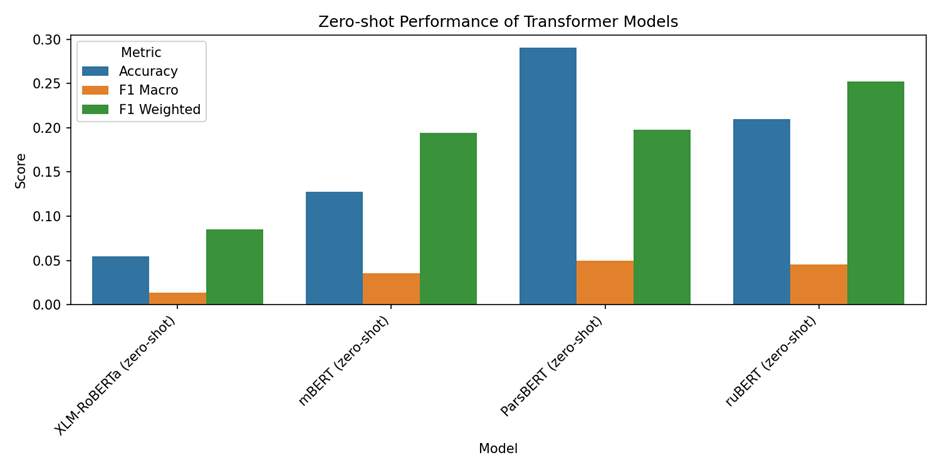}
    \caption{Zero-shot performance of transformer models on the Tajik corpus (without fine-tuning).}
    \label{fig:zeroshot}
\end{figure}

To assess the ability of pre-trained models to generalize to Tajik without targeted fine-tuning, a zero-shot experiment was conducted. The results are presented in Table~\ref{tab:zeroshot} and Figure~\ref{fig:zeroshot}. The data are taken from \texttt{zeroshot\_results.csv}.

\begin{table}[htbp]
\centering
\caption{Zero-shot classification results (without fine-tuning)}
\label{tab:zeroshot}
\begin{tabular}{lccc}
\toprule
\textbf{Model (zero-shot)} & \textbf{Accuracy} & \textbf{Macro F1} & \textbf{Weighted F1} \\
\midrule
XLM-RoBERTa-large & 0.055 & 0.014 & 0.085 \\
mBERT & 0.128 & 0.035 & 0.194 \\
ParsBERT & 0.290 & 0.049 & 0.197 \\
ruBERT & 0.210 & 0.045 & 0.252 \\
\bottomrule
\end{tabular}
\end{table}

All models in the zero-shot setting exhibit quality close to random guessing (with 12 equiprobable classes, random accuracy would be approximately 0.083). The highest accuracy (0.290) is shown by ParsBERT—a model specialized for Persian—which empirically confirms the hypothesis of high typological and lexical proximity between Tajik and Persian. Notably, ruBERT achieves the highest weighted F1 (0.252), which may be attributed both to the shared Cyrillic script and to certain structural similarities in the part-of-speech system resulting from prolonged language contact.

The visualization of zero-shot metrics in Figure~\ref{fig:zeroshot} highlights an asymmetry in the quality of pre-trained representations: ParsBERT leads in accuracy, whereas ruBERT shows a better balance due to weighted F1-score. The critically low level of macro F1 ($< 0.05$) across all models indicates that without parameter-efficient fine-tuning, transformers are incapable of isolating Tajik morphology from the general multilingual latent space. The absolute zero-shot metric values remain unsatisfactory for practical application, underscoring the necessity of a fine-tuning stage on target data even for the task of isolated word classification.

\subsection{Per-Class Analysis}
\begin{figure}[htbp]
    \centering
    \includegraphics[width=0.8\linewidth]{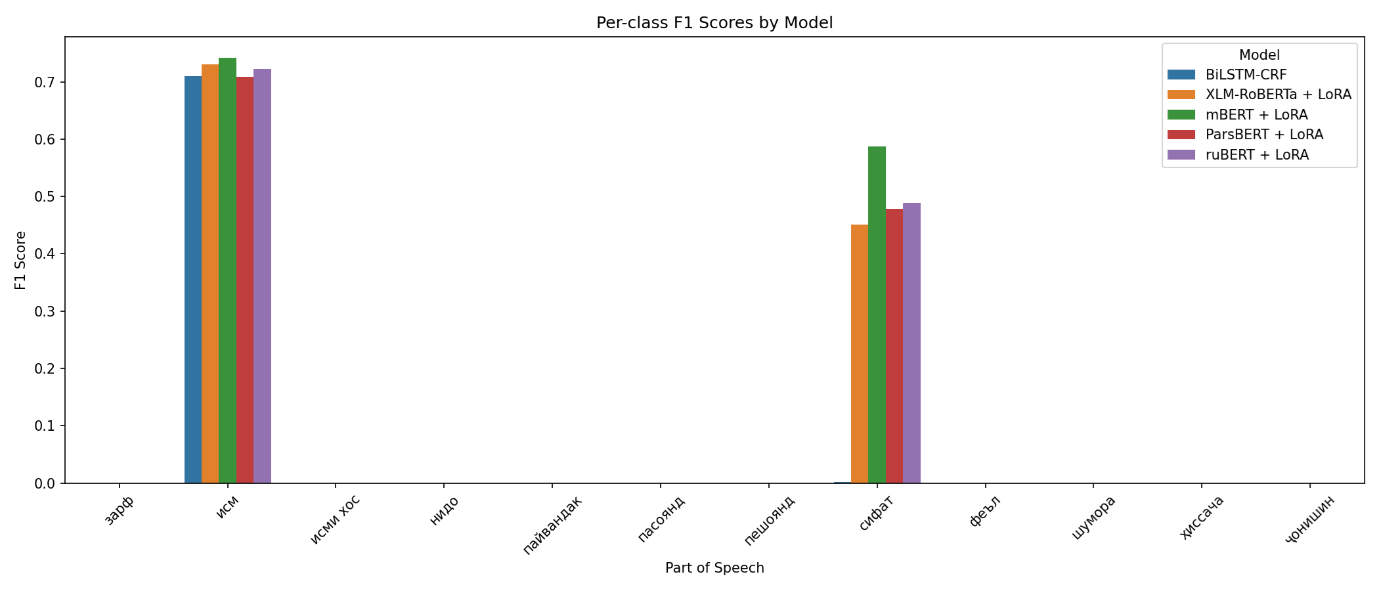}
    \caption{Distribution of per-class F1-scores across part-of-speech categories for all compared architectures.}
    \label{fig:per_class}
\end{figure}

To gain a detailed understanding of the strengths and weaknesses of the models, an analysis of F1-score was conducted for each of the 12 part-of-speech classes based on data from \texttt{per\_class\_f1.csv}. The corresponding values are summarized in Table~\ref{tab:per_class} and visualized in Figure~\ref{fig:per_class}.

\begin{table}[htbp]
\centering
\caption{Per-class F1-scores on the test set}
\label{tab:per_class}
\small
\begin{tabular}{lccccc}
\toprule
\textbf{Class (POS)} & \textbf{BiLSTM-CRF} & \textbf{XLM-R + LoRA} & \textbf{mBERT + LoRA} & \textbf{ParsBERT + LoRA} & \textbf{ruBERT + LoRA} \\
\midrule
{\fontencoding{T2A}\selectfont исм} (noun) & 0.710 & 0.730 & 0.742 & 0.709 & 0.722 \\
{\fontencoding{T2A}\selectfont сифат} (adj.) & 0.001 & 0.451 & 0.587 & 0.479 & 0.489 \\
{\fontencoding{T2A}\selectfont зарф} (adv.) & 0.0 & 0.0 & 0.0 & 0.0 & 0.0 \\
{\fontencoding{T2A}\selectfont феъл} (verb) & 0.0 & 0.0 & 0.0 & 0.0 & 0.0 \\
{\fontencoding{T2A}\selectfont исми хос} (prop.n.) & 0.0 & 0.0 & 0.0 & 0.0 & 0.0 \\
{\fontencoding{T2A}\selectfont нидо} (interj.) & 0.0 & 0.0 & 0.0 & 0.0 & 0.0 \\
{\fontencoding{T2A}\selectfont шумора} (num.) & 0.0 & 0.0 & 0.0 & 0.0 & 0.0 \\
{\fontencoding{T2A}\selectfont пайвандак} (conj.) & 0.0 & 0.0 & 0.0 & 0.0 & 0.0 \\
{\fontencoding{T2A}\selectfont ҷонишин} (pron.) & 0.0 & 0.0 & 0.0 & 0.0 & 0.0 \\
{\fontencoding{T2A}\selectfont ҳиссача} (part.) & 0.0 & 0.0 & 0.0 & 0.0 & 0.0 \\
{\fontencoding{T2A}\selectfont пешоянд} (prep.) & 0.0 & 0.0 & 0.0 & 0.0 & 0.0 \\
{\fontencoding{T2A}\selectfont пасоянд} (postp.) & 0.0 & 0.0 & 0.0 & 0.0 & 0.0 \\
\bottomrule
\end{tabular}
\end{table}

Analysis of Table~\ref{tab:per_class} reveals an extremely limited ability of all models to recognize parts of speech beyond the two dominant classes—nouns ({\fontencoding{T2A}\selectfont исм}) and adjectives ({\fontencoding{T2A}\selectfont сифат}). For the noun class, all models achieve relatively high F1-scores (0.71--0.74), which is explained by its absolute predominance in the training set (55.2\%). For the adjective class, a significant spread is observed: from 0.001 for BiLSTM-CRF to 0.587 for mBERT + LoRA. The remaining ten classes, whose combined share in the corpus is less than 10\%, are not predicted by any of the models (F1 = 0.0).

As Figure~\ref{fig:per_class} clearly demonstrates, all models exhibit practically binary behavior: high effectiveness on the class ``{\fontencoding{T2A}\selectfont исм}'' (F1 $\approx$ 0.71--0.74) and moderate on ``{\fontencoding{T2A}\selectfont сифат}'' (F1 $\approx$ 0.45--0.59), with a complete lack of predictive power for the other ten categories. Notably, mBERT + LoRA shows the most balanced profile, whereas BiLSTM-CRF practically fails to identify adjectives (F1 $\approx$ 0.001). Zero values for function words are linguistically motivated by their homonymy with content words in isolated form. For instance, words like ``{\fontencoding{T2A}\selectfont ки}'' or ``{\fontencoding{T2A}\selectfont ҳар}'' cannot be unambiguously assigned to a specific part-of-speech class without syntactic context, making the task unsolvable for context-independent classifiers regardless of their architectural complexity.

\subsection{Error Analysis and Confusion Matrix}
\begin{figure}[htbp]
    \centering
    \includegraphics[width=0.8\linewidth]{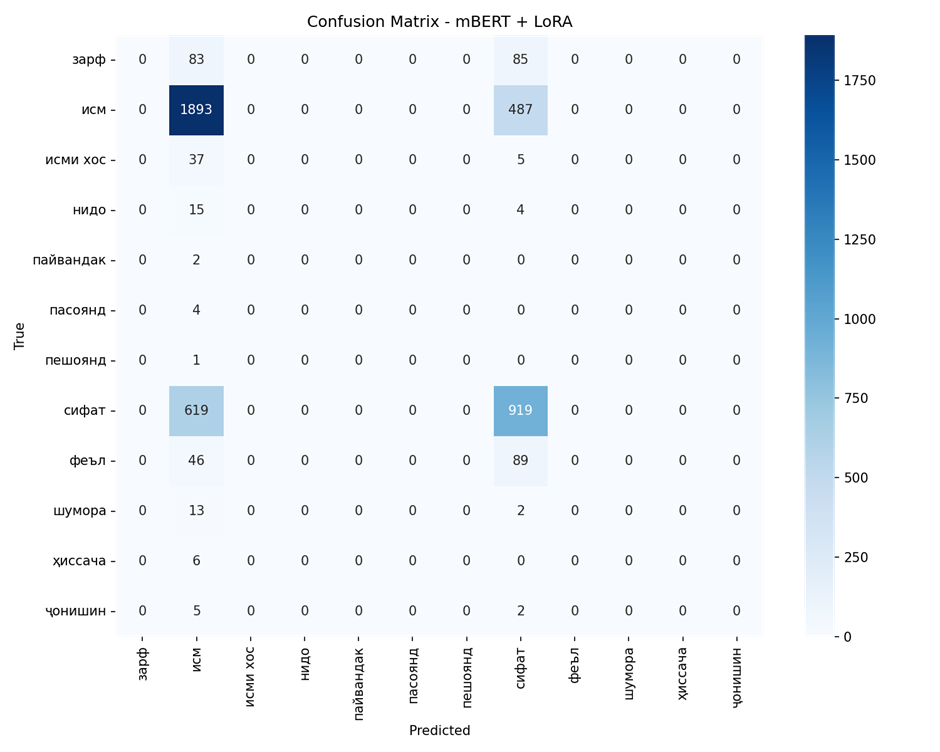}
    \caption{Confusion matrix for the mBERT + LoRA model: distribution of true and predicted part-of-speech labels on the test set.}
    \label{fig:confusion}
\end{figure}

For the best-performing model—mBERT + LoRA—a confusion matrix was constructed, visualizing the distribution of predictions relative to the true labels, as shown in Figure~\ref{fig:confusion}.

Analysis of Figure~\ref{fig:confusion} confirms that the bulk of mispredictions is concentrated in the block of mutual confusion between the classes ``{\fontencoding{T2A}\selectfont исм}'' and ``{\fontencoding{T2A}\selectfont сифат}''. The values 1893 (correctly classified nouns) and 919 (correctly classified adjectives) contrast with the off-diagonal elements 619 and 487, indicating systematic interchange between these categories. This error has a deep morphological basis: in Tajik, adjectives in attributive position often coincide in form with nouns, and their distinction requires analysis of izafet constructions or syntactic position, which is unavailable in the current formulation. The remaining classes are either not predicted at all (zero columns/rows) or erroneously absorbed by the dominant categories, consistent with the previously described limitations of zero-shot and per-class analysis.

Thus, the confusion matrix clearly demonstrates that in the isolated word classification setting, the fundamental ambiguity between nouns and adjectives remains irreducible, and rare function words are not recognized by any of the examined architectures.

\section{Discussion}
The conducted research represents the first systematic benchmark of automatic part-of-speech tagging for the Tajik language. The obtained results allow us to formulate several fundamental conclusions regarding both the applicability of modern neural network architectures to Tajik linguistic material and the inherent limitations of the POS tagging task in the absence of syntactic context.

First and foremost, the clear superiority of all multilingual transformers adapted with LoRA \citep{hu2022_lora} over the baseline BiLSTM-CRF architecture is striking. The mBERT + LoRA model \citep{devlin2019bert}, which demonstrated the best results (accuracy 0.651, weighted F1 0.618), outperforms BiLSTM-CRF by more than 10 percentage points in accuracy and by nearly 23 points in weighted F1. This result confirms that even without the direct inclusion of the Tajik language in the pre-training corpus, multilingual models are capable of extracting relevant morphological information from word forms by leveraging knowledge acquired from closely related or typologically similar languages. Notably, the specialized Persian model ParsBERT \citep{farahani2021_parsbert}, contrary to expectations, underperformed compared to the universal mBERT. This fact may be explained both by the smaller overall pre-training volume of ParsBERT relative to mBERT and by the absence in the ParsBERT architecture of mechanisms specifically oriented toward script variability, which proves more critical for the Tajik language with its Cyrillic script than lexical proximity to Persian.

The zero-shot evaluation results serve as independent confirmation of the hypothesis concerning the linguistic proximity of Tajik and Persian: the ParsBERT model, without any fine-tuning, achieves an accuracy of 0.290, nearly three times higher than the random baseline. Simultaneously, the relatively high performance of ruBERT \citep{kuratov2019_rubert} (weighted F1 0.252) indicates the significant role of the shared Cyrillic script and, possibly, a certain structural isomorphism in the part-of-speech system that has developed as a result of prolonged language contact. Nevertheless, the absolute values of the zero-shot metrics remain too low for practical use, underscoring the necessity of a fine-tuning stage on the target language even for such a basic task as isolated word classification.

A key and most indicative finding of the study is the dramatic gap between weighted and macro-averaged F1-scores, as well as the complete inability of all models without exception to recognize rare part-of-speech classes (10 out of 12 classes have F1 = 0.0). This phenomenon has a dual nature. On the one hand, it is driven by the critical class imbalance of the training set, in which over 90\% of examples are concentrated in the classes {\fontencoding{T2A}\selectfont исм} (noun) and {\fontencoding{T2A}\selectfont сифат} (adjective). On the other hand, and more fundamentally, in a context-independent classification setting, many function words and even content words cannot be unambiguously assigned to a particular class based on form alone. As demonstrated by the confusion matrix analysis, even within the two dominant classes, there is systematic confusion between nouns and adjectives, which is linguistically motivated by the absence of formal distinctions between them in attributive position and the fundamental impossibility of resolving such ambiguity without syntactic context.

Comparison of the obtained results with data from other low-resource languages is hindered by the lack of direct analogues; however, the achieved accuracy (0.65) and weighted F1 (0.62) values align with the typical range of indicators for initial works in the field of POS tagging for languages with a shortage of annotated corpora. Thus, the present study not only introduces the first benchmark for the Tajik language into scientific discourse but also establishes a methodological framework for subsequent work, fixing a lower bound on the quality achievable without the use of syntactic context.

\section{Conclusion}
In this work, the first systematic benchmark of automatic part-of-speech tagging for the Tajik language was performed using the TajPersParallel parallel lexical corpus \citep{arabov2026_tajperslexon}. A comparison was conducted of five neural network architectures: the classical BiLSTM-CRF model and four multilingual transformers (XLM-RoBERTa-large \citep{conneau2020_xlmr}, mBERT \citep{devlin2019bert}, ParsBERT \citep{farahani2021_parsbert}, and ruBERT \citep{kuratov2019_rubert}), adapted using the parameter-efficient fine-tuning method LoRA \citep{hu2022_lora}.

It was experimentally established that the best performance is achieved by the mBERT + LoRA model, which attains an accuracy of 0.651, a macro-averaged F1-score of 0.111, and a weighted F1-score of 0.618 on the test set. It was shown that in the absence of syntactic context, all considered models face fundamental difficulties in resolving morphological ambiguity, successfully classifying only the two most frequent classes ({\fontencoding{T2A}\selectfont исм} and {\fontencoding{T2A}\selectfont сифат}) while demonstrating zero effectiveness for rare function words. Zero-shot evaluation confirmed the greatest typological proximity of Tajik to Persian and also revealed the significant role of the shared Cyrillic script and Russian as an intermediary language.

The scientific significance of this work lies in the creation of the first reproducible reference benchmark for the task of Tajik POS tagging, as well as in the publication of pre-trained models and the source code of the experiments in open access, thereby enabling independent verification of the results and their utilization in applied developments.

Among the objective limitations of the present study is the training of models exclusively on isolated lexical units, which precludes the consideration of syntactic context and limits the practical applicability of the obtained POS taggers for the analysis of connected texts. Prospects for future work are primarily associated with expanding the TajPersParallel corpus to include annotated sentences, which will enable a transition to full-fledged contextual POS tagging. Additionally, relevant directions include the application of data augmentation methods and weighted loss functions to address class imbalance, investigation of the influence of various LoRA adaptation schemes, and the integration of the resulting models into specialized NLP libraries for the Tajik language.

\bibliographystyle{unsrtnat}
\bibliography{references}  

\end{document}